%% file: main.tex
\documentclass[11pt,letterpaper,logo,dvipsnames,table]{thuair}

\usepackage{hyperref}
\usepackage{url}
\usepackage{graphicx}
\usepackage{xspace}
\usepackage{enumitem} % Add this package to your preamble
\usepackage{lipsum} % For dummy text to show alignment
\usepackage{lipsum} % For dummy text to show alignment
\usepackage{tcolorbox}
\usepackage{natbib}
\usepackage{caption}
\usepackage{subcaption}

\newtcolorbox{AIbox}[1]{colback=gray!10!white,colframe=black!80!white,title=#1}

\usepackage{booktabs}
\usepackage{multirow}
\usepackage{tabularx} % 或者 array, makecell
\usepackage{calc}

\usepackage{array}
\newcolumntype{C}[1]{>{\centering\let\newline\\\arraybackslash\hspace{0pt}}p{#1}}

\hypersetup{
    colorlinks = true,
    citecolor = {blue},
}

\newcommand{\name}{ParaThinker\xspace} % name of the tool
\newcommand{\ie}{\textit{i}.\textit{e}.,~}
\newcommand{\eg}{\textit{e}.\textit{g}.,~}
\title{\name: Native Parallel Thinking as a New Paradigm to Scale LLM Test-time Compute}
\author{
Hao Wen$^{1,*}$, Yifan Su$^{1,*,\ddagger}$, Feifei Zhang$^{1}$, Yunxin Liu$^{1}$, Yunhao Liu$^{2}$, \mbox{Ya-Qin Zhang}$^{1}$, Yuanchun Li$^{1,\dagger}$ \\
$^{1}$Institute for AI Industry Research (AIR), Tsinghua University \\
$^{2}$Global Innovation Exchange \& Department of Automation, Tsinghua University \\~\\
$^*$Equal contribution. \\
$^\dagger$Corresponding author: Yuanchun Li (liyuanchun@air.tsinghua.edu.cn). \\
$^\ddagger$Work done during internships at AIR, Tsinghua University.\\
\textbf{Source Code}: \url{https://github.com/MobileLLM/ParaThinker} \\
}
\begin{abstract}
Recent advances in Large Language Models (LLMs) have been driven by test-time compute scaling - a strategy that improves reasoning by generating longer, sequential thought processes. 
While effective, this approach encounters a significant bottleneck as computation increases, where further computation offers only marginal performance gains. 
We argue this ceiling is not an inherent limit of the model's capability but a flaw in the scaling strategy itself, a phenomenon we term \emph{``Tunnel Vision''}, where a model's imperfect initial steps lock it into a suboptimal reasoning path. 
To overcome this, we introduce a new scaling paradigm: \emph{native thought parallelism}. We present \emph{\name}, an end-to-end framework that trains an LLM to generate multiple, diverse reasoning paths in parallel and synthesize them into a superior final answer. 
By exploring different lines of thoughts simultaneously, \emph{\name} effectively sidesteps the \emph{Tunnel Vision} issue and unlocks the model's latent reasoning potential. 
Our approach demonstrates that scaling compute in parallel (width) is a more effective and efficient way to superior reasoning than simply scaling sequentially (depth). 
On challenging reasoning benchmarks, \emph{\name} achieves substantial accuracy improvements over sequential LLMs (12.3\% for 1.5B and 7.5\% for 7B models on average with 8 parallel paths), while adding only negligible latency overhead (7.1\%). 
This enables smaller models to surpass much larger counterparts and establishes parallel thinking as a critical, efficient dimension for scaling future LLMs.

\end{abstract}

\begin{document}

\maketitle

\begin{figure*}[htbp]
  \centering
  \includegraphics[width=0.9\linewidth]{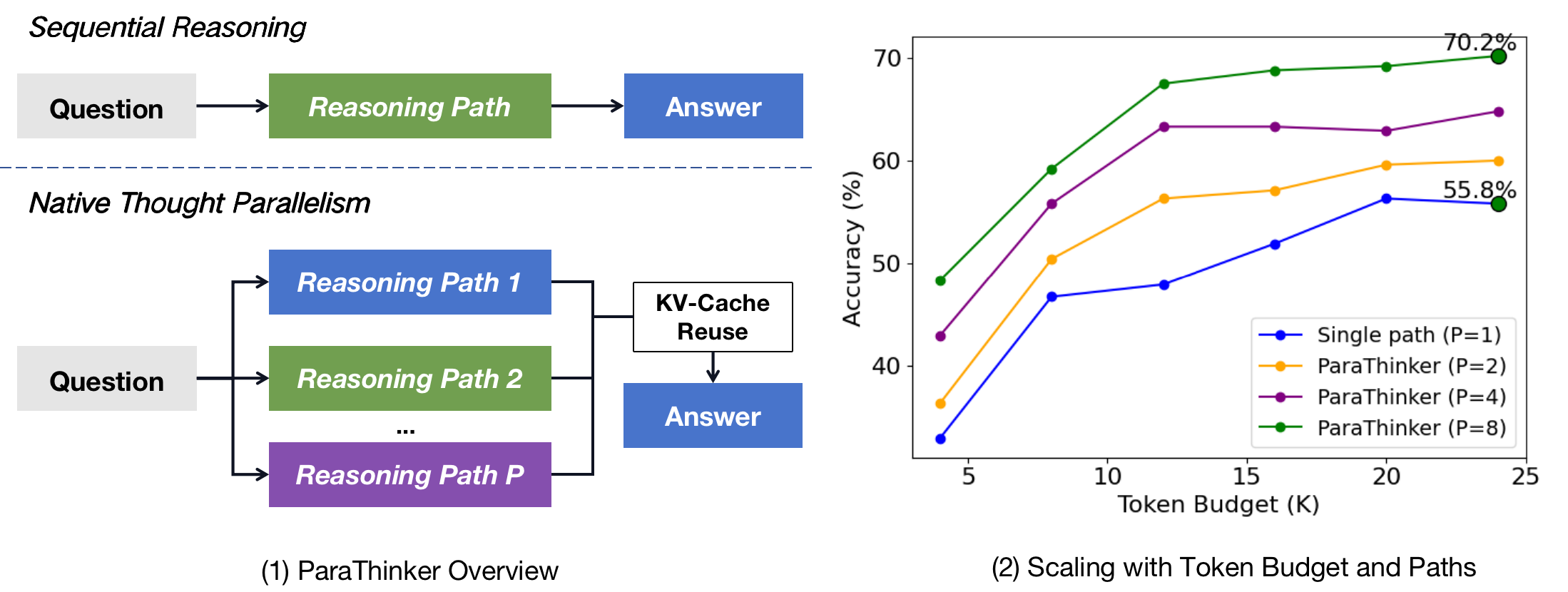}
  % \vspace{-8mm}
  \caption{Sequential vs. Parallel reasoning with ParaThinker framework and scaling results. (1) Illustrations of \name. (2) Parallel scaling results of \name-7B on AIME 2024 with varying numbers of reasoning paths ($P$). ``Token budget'' refers to the maximum token length allowed per reasoning path.}
  \label{fig:main}
\end{figure*}

\input{tex/intro}
\input{tex/motivation}

\input{tex/method}

\input{tex/experiment}

\input{tex/related}
\input{tex/conclusion}

\bibliography{iclr2025_conference}
\bibliographystyle{iclr2025_conference}

\input{tex/appendix}

% \appendix
% \section{Appendix}

\end{document}

%% file: tex/intro.tex
\section{Introduction}

The remarkable progress of Large Language Models (LLMs) has been largely driven by the principle of scaling. This evolution began with pretraining compute scaling and has recently shifted to post-training or test-time compute scaling. Notable examples of test-time scaling, such as OpenAI o1 \citep{OpenAI2024o1} and DeepSeek-R1 \citep{deepseekr1}, have demonstrated that training the models to ``think longer'' (\ie decode more tokens before generating the final answer) can unlock superior reasoning abilities for complex problems \citep{yang2025qwen3technicalreport, kimiteam2025kimik15scalingreinforcement, snell2024scalingllmtesttimecompute, wu2025inferencescalinglawsempirical}. 

However, extending test-time compute does not lead to constant performance improvement in today's reasoning LLMs, where accuracy improvements diminish and eventually stagnate after a certain number of decoding steps. 
% We call this phenomenon ``test-time scaling bottleneck'', \ie accuracy improvements diminish and eventually stagnate after a certain number of decoding steps. 
This has fueled discussions around ``LLM overthinking'' \citep{ghosal2025doesthinkinghelpunderstanding, chen2025think23overthinkingo1like, fan2025missingpremiseexacerbatesoverthinking, cuadron2025dangeroverthinkingexaminingreasoningaction, li2025selfbudgeter}, where models expend excessive computation on problems, with the additional reasoning steps yielding minimal or no benefit to the final answer.

In this paper, we investigate the problem of test-time scaling bottleneck by raising a fundamental question: \emph{Does the test-time scaling bottleneck stem from the inherent limitations of the model's capability, or from the imperfect test-time compute strategy?} 
The answer to this question is important for understanding the bottleneck of test-time scaling. Our findings reveal that, given a fixed decoding token budget, the conventional self-refinement reasoning paradigm (adopted in state-of-the-art reasoning models like o1 and R1) constantly converges at a low accuracy that can be achieved with other simple scaling strategies (\eg majority voting) under the same token budget. This suggests that the model's underlying capability is not the primary bottleneck; rather, the way we orchestrate test-time compute can be improved.

Through a closer analysis of the reasoning process in LLMs, we find that the reasoning performance is often constrained by the model's initial thoughts, a phenomenon we refer to as \emph{Tunnel Vision}. Specifically, the first few tokens generated in a Chain-of-Thought (CoT) can lock the model into a suboptimal reasoning path, preventing it from discovering more effective ideas in subsequent decoding steps.

% For example, when solving a quadratic equation, if the model mistakenly expands $(x+1)^2$ as $x^2+1$, the subsequent reasoning remains locked into this flawed expansion and cannot recover the correct solution. \lyc{Example}

Based on these insights, we argue that the reasoning process of LLMs should be executed in a parallel, multi-threaded manner. By ensuring each thinking thread operates independently, we can mitigate Tunnel Vision and foster a diversity of thought. 
Furthermore, parallel thinking offers significant deployment advantages, as the decoding process can be batched to better utilize memory bandwidth, which in turn leads to improved arithmetic intensity \citep{roofline} (the ratio of floating-point operations to total data movement).

% Majority voting is a straightforward implementation of parallel thinking and has proven effective at surpassing the test-time scaling bottleneck for problems with discrete answers. However, its applicability is limited. An alternative, aggregating diverse thoughts by re-prefilling them into the model, introduces substan tial computational overhead. Crucially, both methods depend on the model's inherent stochasticity to generate diverse outputs, a property that is neither guaranteed nor controllable. 
% \wh{I have changed the following paragraph to include more details}
To put parallel thinking into practice, we introduce an end-to-end solution, ParaThinker, 
which enables native parallel thinking in LLMs by allowing the model to generate diverse thoughts and aggregate them into a final answer. 
The major challenges to developing ParaThinker include how to induce thought diversity and how to avoid thought conflict, which we address by introducing trainable control tokens to trigger distinct reasoning trajectories, thought-specific positional embeddings to distinguish different paths, and a two-phase attention mask design that enforces independence during reasoning and controlled integration during summarization. 
Specifically, our solution features three core innovations:

\begin{itemize}[leftmargin=*, itemsep=0pt, parsep=0pt, topsep=2pt]  % This option removes the indentation
    \item \textbf{Specialized Control Tokens: }We introduce a set of trainable tokens (\eg \texttt{<think i>}) to explicitly guide the model’s generation. Each \texttt{<think i>} token prompts the model to initiate a distinct reasoning path, which ensures diversity in reasoning.
    \item \textbf{Thought-Specific Positional Embedding: }To resolve positional ambiguity when merging parallel thoughts, we augment the standard positional encoding with a unique, learnable embedding for each reasoning path. This allows the model to unambiguously differentiate the origin of each token during the final summarization stage.
    \item \textbf{SFT Training Pipeline: }We employ a scalable supervised fine-tuning (SFT) strategy where the model is trained on reasoning paths sampled from a teacher model. By randomly assigning the specialized \texttt{<think i>} tokens during this process, the model learns to generalize, enabling it to generate more parallel paths at inference time than were seen during training.
\end{itemize}

We evaluate ParaThinker on challenging reasoning benchmarks: AIME 2024, AIME 2025, AMC 2023, and MATH-500 \citep{math500}, against baselines such as standard autoregressive reasoning \citep{deepseekr1}, majority voting \citep{majority_voting}, and re-prefilling. Our approach demonstrates a remarkable leap in performance, achieving significantly improved accuracy with slight latency overhead. This efficiency allows smaller LLMs equipped with our native thought parallelism to outperform much larger, standard reasoning models, charting a new path for scaling test-time compute.

In summary, the contributions of our work are: (1) We characterize the test-time scaling bottleneck in LLM reasoning and attribute it to a narrow reasoning pathway, termed Tunnel Vision, which restricts the model’s exploration during generation. (2) We propose and demonstrate that thought parallelism is a better way to scale LLM test-time compute. (3) We introduce an end-to-end solution to enable native parallel thinking.
% The resulting model, \name, achieves new state-of-the-art accuracy on common LLM reasoning benchmarks with remarkable efficiency. \syf{Remove sota here}
The resulting model, \name, achieves higher accuracy than sequential LLMs by 12.3\% and 7.5\% for 1.5B and 7B models, respectively. Compared with majority voting, \name further improves accuracy by 4.3\% and 2.0\%. 

%% file: tex/motivation.tex
\section{Understanding the Scaling Bottleneck}

\begin{figure}[htbp]
    \centering
    
    \begin{subfigure}{0.32\textwidth}
        \centering
        \includegraphics[width=\linewidth]{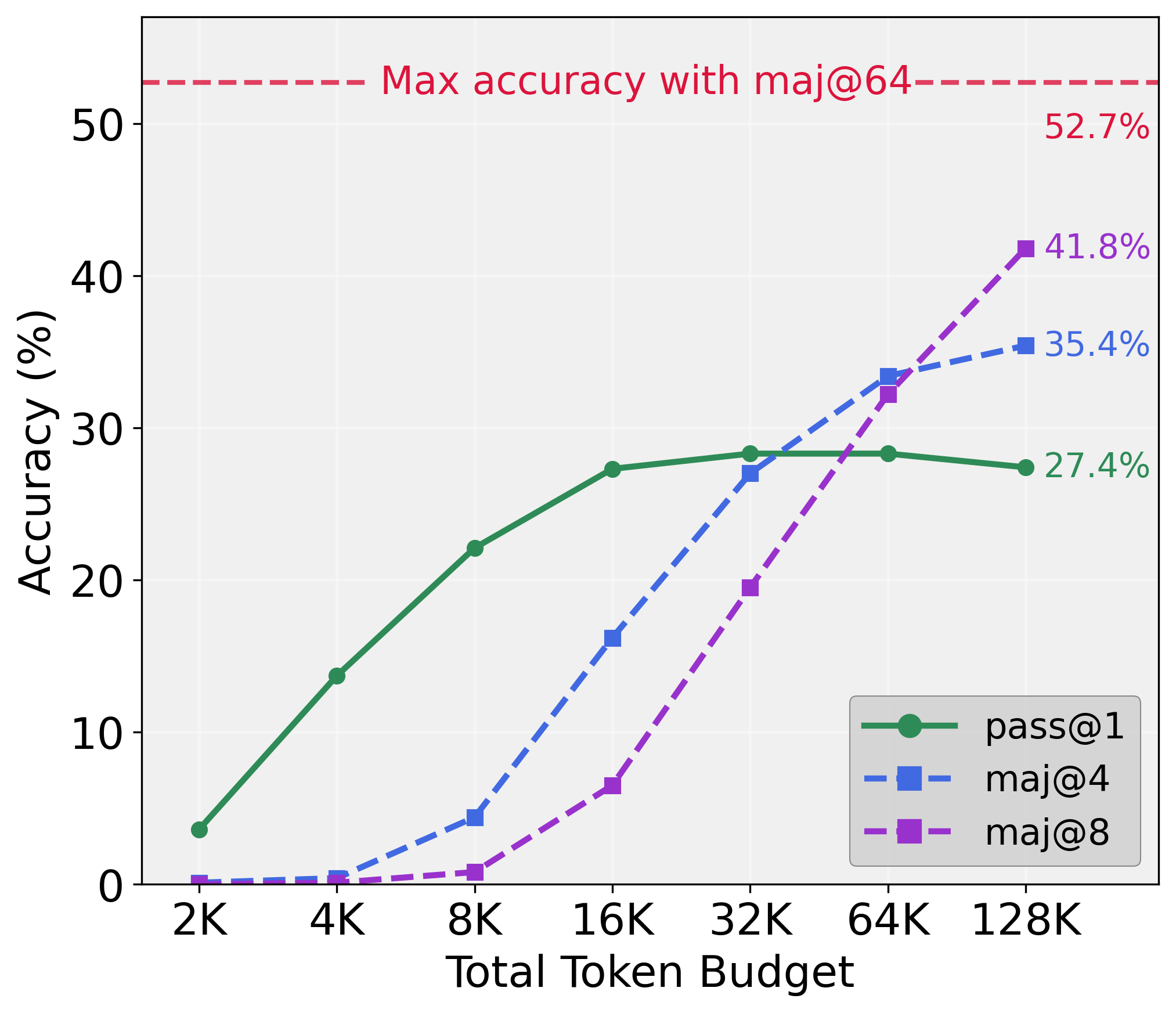}
        \caption{Scaling Bottleneck}
        \label{fig:scaling_bottleneck}
    \end{subfigure}
    \hfill
    \begin{subfigure}{0.32\textwidth}
        \centering
        \includegraphics[width=\linewidth]{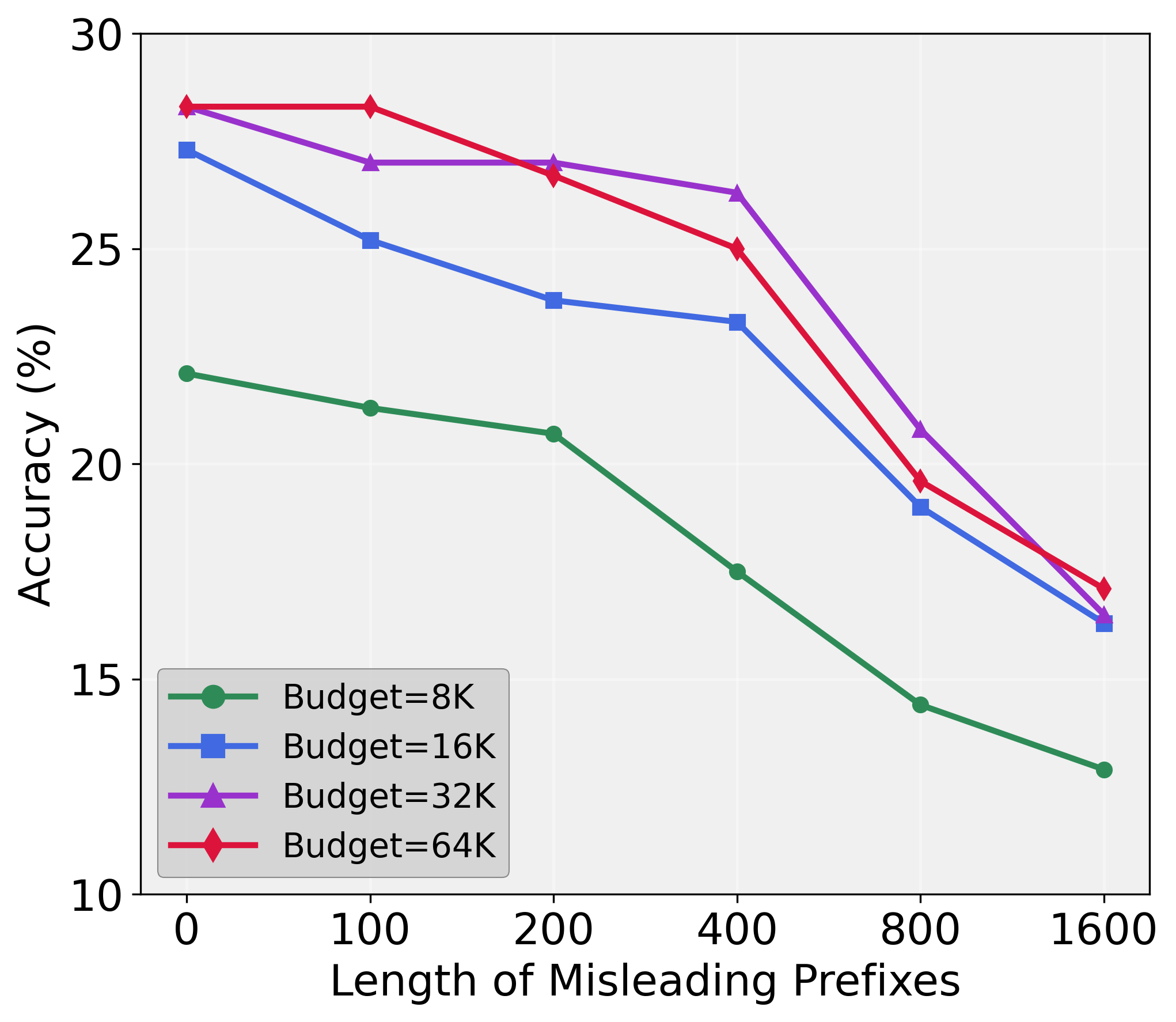}
        \caption{Tunnel Vision}
        \label{fig:tunnel_vision}
    \end{subfigure}
    \hfill
    \begin{subfigure}{0.32\textwidth}
        \centering
        \includegraphics[width=\linewidth]{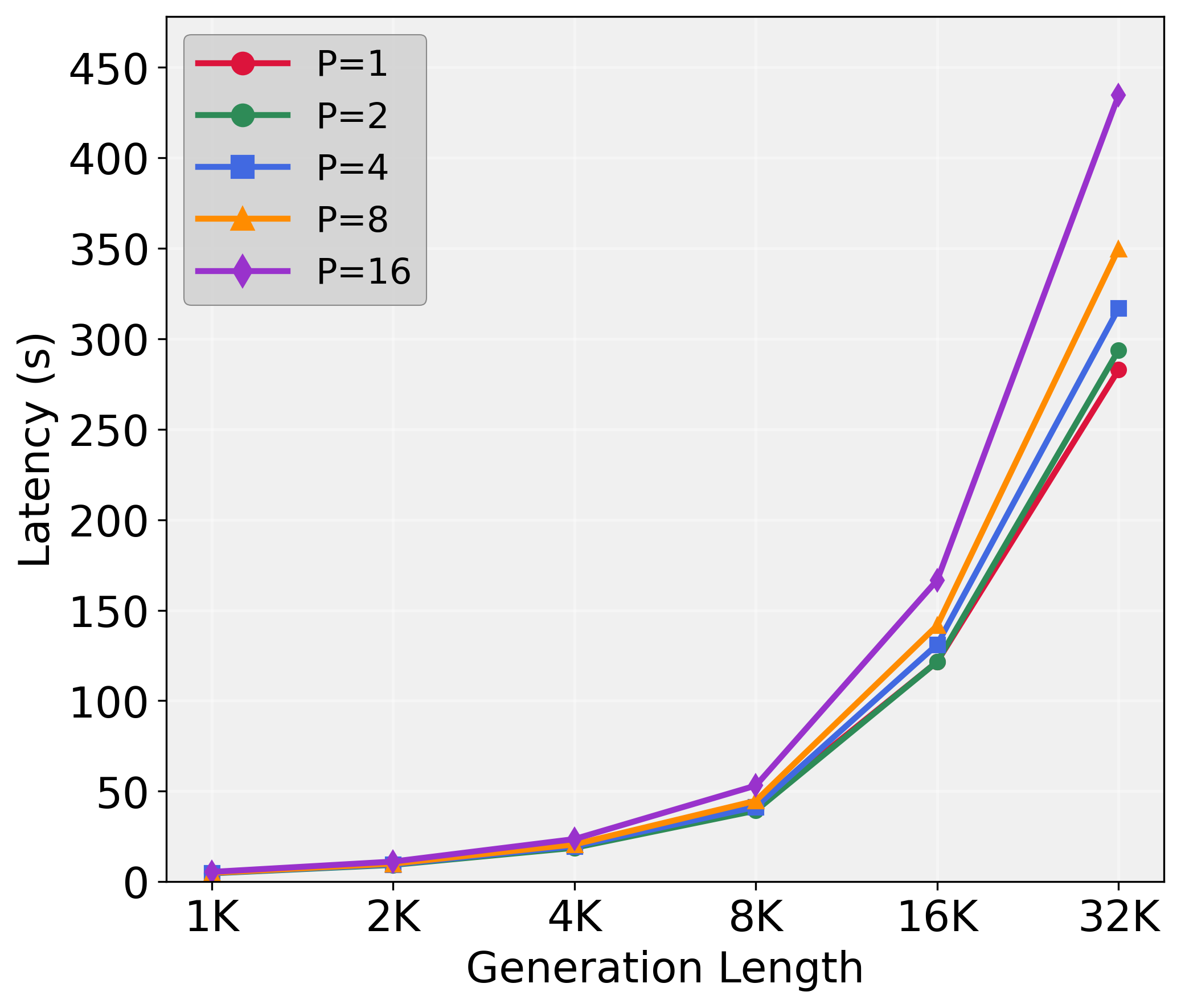}
        \caption{Parallel Decoding Efficiency}
        \label{fig:parallel_decoding}
    \end{subfigure}
    
    \caption{
        Diagnosing the limitations of sequential reasoning and the potential of parallelism.
        All experiments use DeepSeek-R1-Distill-Qwen-1.5B on the AIME 2024 benchmark. 
        (a) Scaling Bottleneck: Accuracy against the total number of token budget (for majority voting \eg maj@4, the total token budget is the sum across all parallel paths).
        (b) Tunnel Vision: Ability to recover from its own potential mistakes with different lengths of misleading prefixes. The model generates solutions starting from flawed prefixes of length $n\in\{0,100,\dots,1600\}$, denoting the first $n$ tokens of reasoning paths from the same model that previously resulted in a wrong answer.
        (c) Parallel Decoding Efficiency:  Latency taken to decode $P\in\{1, 2,4,8,16\}$ parallel paths, each of length $n\in\{1K,\dots,32K\}$.
    }
    \label{fig:combined_figures}
\end{figure}

To empirically ground our investigation, we first characterize the limitations of conventional test-time scaling. 
We identify a test-time scaling bottleneck that suggests a fundamental inefficiency in sequential reasoning (Section~\ref{sec:plateau}). Then, we diagnose its underlying cause, a phenomenon we term Tunnel Vision (Section~\ref{sec:trap}). 
Then, we propose parallel scaling as an effective and efficient solution to it (Section~\ref{sec:motivation_parallel}). 

\subsection{Is the Bottleneck Due to Limited LLM Capability or Suboptimal Scaling Strategy? }
\label{sec:plateau}

We examine the performance bottleneck in reasoning LLMs by evaluating the DeepSeek-R1-distill-Qwen-1.5B model \citep{deepseekr1} on the AIME 2024 benchmark under various computational budgets. 
We control the budget by imposing a per-response token limit $B$ on the reasoning path. If the model fails to terminate naturally, we truncate the output and force termination by appending a terminal token (\texttt{</think>}). 
We also evaluate majority voting \citep{majority_voting} over $P\in\{4,8,16, 32, 64\}$ parallel samples, with each sample allocated $B/P$ tokens. 
For clarity, we plot only $P=4$ and $P=8$ results in the accuracy–budget curves, and report the maximum accuracy obtained with $P=64$. 
The results, shown in Figure \ref{fig:scaling_bottleneck}, demonstrate that the performance of a single reasoning path (green line) quickly meets bottleneck, with additional tokens yielding negligible gains.

% The results confirm for single-sequence test-time scaling, once the bottleneck is reached, allocating more tokens yields negligible gains. 
While some recent works attribute this phenomenon to ``LLM overthinking'' and attempt to solve it by compressing the model's output for more concise reasoning, these compressed models still encounter a bottleneck \citep{Sun2025TimesUp, chen2025think23overthinkingo1like, fan2025missingpremiseexacerbatesoverthinking, cuadron2025dangeroverthinkingexaminingreasoningaction}. 
Different from these approaches, we investigate the fundamental cause of this bottleneck to determine how to further bootstrap LLM test-time scaling. 
And results in Figure \ref{fig:scaling_bottleneck} shows that majority voting can break through this bottleneck under the same total token budget, and the majority@64 with 2,048K total token budget (32K for each reasoning path), achieves a final accuracy far higher than the single-path approach. 
This significant gap suggests that \textit{the bottleneck is not a hard limit of the model’s reasoning capacity, but rather a symptom of the suboptimal test-time scaling strategy. }
Simply allocating more test-time compute to a single-sequence LLM is not as effective as exploring multiple reasoning paths.

\subsection{The Tunnel Vision of Sequential Test-Time Scaling}
\label{sec:trap}

We hypothesize the bottleneck arises because an LLM's early token choices irreversibly commit it to a specific line of thought, making it difficult to escape initial errors. We call this Tunnel Vision: flawed initial reasoning locks the model into a suboptimal trajectory from which it cannot recover. % Since an LLM's output is conditioned on both the LLM's parameter and all preceding tokens, and the model can achieve a correct result after repeated parallel trials, we ask: \textbf{are LLMs misled into wrong answers by their own initial token samples?}
To test this, we investigate the model's \textit{recovery capacity} from erroneous starting points:
For each AIME 2024 problem, we use DeepSeek-R1-Distill-Qwen-1.5B \citep{deepseekr1} to generate multiple samples.  
From the samples that produce incorrect answers, we extract prefixes of its flawed reasoning at lengths of 0, 100, 200, 400, 800, and 1600 tokens. 
We then prompt the model to continue generating from these erroneous prefixes and measure its final accuracy by sampling 16 times and calculating the average accuracy.  
The results, plotted in Figure \ref{fig:tunnel_vision}, show a clear negative correlation: the longer the erroneous prefix, the lower the final accuracy. This indicates that \textit{the scaling bottleneck is a direct symptom of Tunnel Vision, where flawed initial tokens lock the model into a suboptimal reasoning path.} The longer the flawed prefix, the harder it is for the model to pivot to a correct solution, even with ample remaining budget. 

\subsection{Native Thought Parallelism: An Effective and Efficient Solution}
\label{sec:motivation_parallel}

Given the limitations of sequential reasoning exposed by Tunnel Vision, exploring multiple lines of thought in parallel is a promising solution. While methods like majority voting \citep{majority_voting} validate the benefit of parallelism, they are primarily applicable to tasks with easily quantifiable or verifiable outputs (\eg multiple-choice questions, numerical answers). Their utility is limited for more complex, open-ended domains (\eg complex agentic workflows, coding, or mathematical proofs).

This necessitates a native parallel framework—one that enables an LLM to generate and integrate multiple reasoning paths within a single, end-to-end forward pass. Such a system should scale with the number of parallel paths while maintaining high computational efficiency. 
Previous approaches to parallel computation typically rely on external verifiers for search \citep{snell2024scalingllmtesttimecompute, ghosal2025doesthinkinghelpunderstanding}, which introduces a scalability bottleneck. Recent methods such as architectural parallelism \citep{chen2025parallelscalinglawlanguage}, still generate tokens sequentially, leaving them vulnerable to the same phenomenon, Tunnel Vision.

The efficiency of native thought parallelism stems from its hardware-friendly nature. 
The primary bottleneck in LLM decoding speed is typically memory access (parameter load, KV load/store), not raw computation \citep{magicdec}. By generating $P$ parallel reasoning paths simultaneously, we increase the computational workload for each memory access, improving the arithmetic intensity and better utilizing the GPU's processing power. 
We test the efficiency of parallel generation in Figure \ref{fig:parallel_decoding}, where we record the inference time of DeepSeek-R1-Distill-Qwen-1.5B based on vLLM framework \citep{vllm} on a single A800 GPU with different generation lengths, using a batch size of one. 
For a small number of parallel paths, generating $L$ tokens for every path takes nearly the same amount of time as generating $L$ tokens for a single path. 
Remarkably, even when decoding 16 parallel paths of length $L$, the total latency is less than $2\times$ of decoding a single path of the same length. 
This hardware-level efficiency makes parallel exploration a practical and scalable strategy for overcoming Tunnel Vision and unlocking superior reasoning performance.

%% file: tex/method.tex
\section{Model Design}
\label{sec:approach}

\begin{figure*}[htbp]
  \centering
  \includegraphics[width=0.95\linewidth]{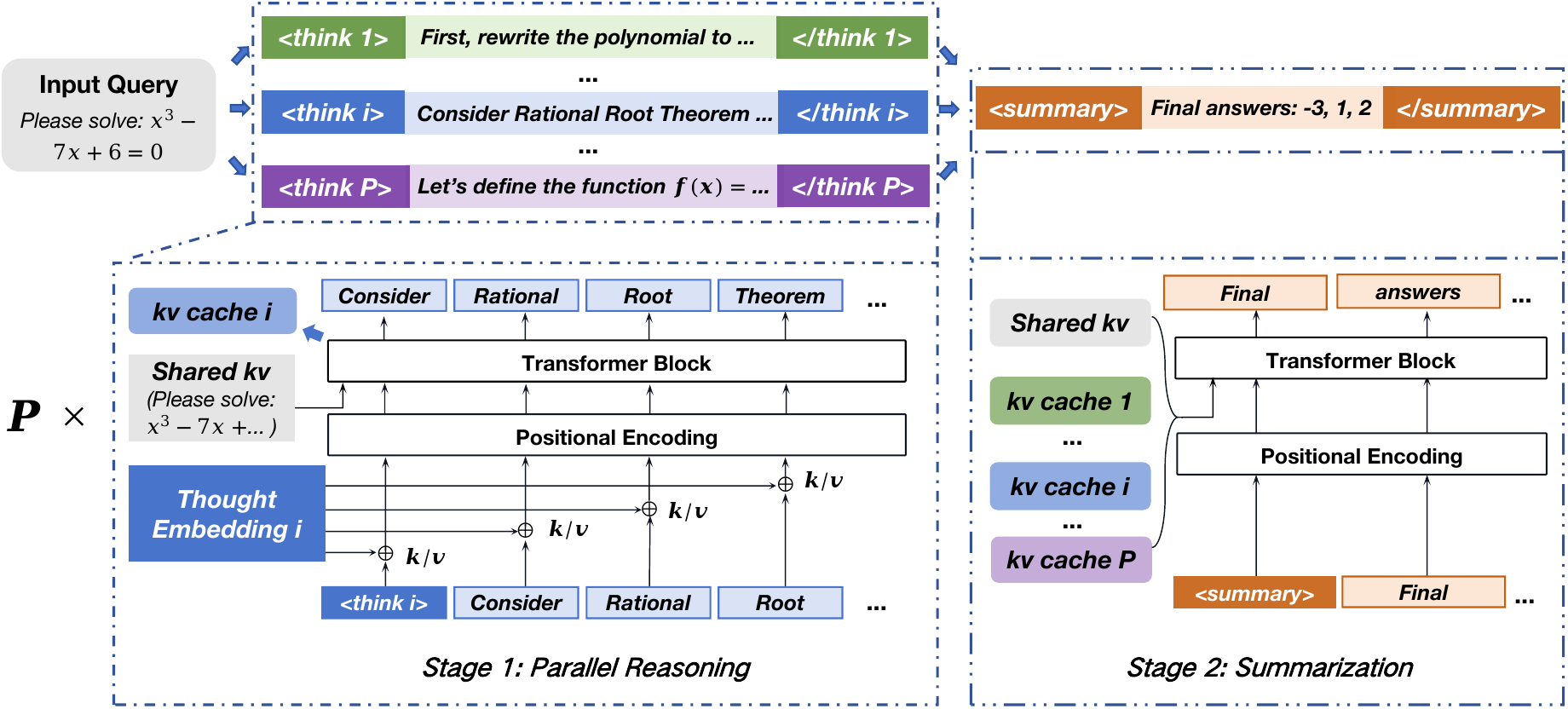}
  % \vspace{-8mm}
  \caption{\name architecture. For an input question, \name processes it in two stages: (1) Parallel Reasoning: \name generates $P$ reasoning paths in parallel, guided by special \texttt{<think i>} tokens to encourage diverse thoughts, and employs thought embeddings to separate paths and prevent positional ambiguity; (2) Summarization: \name merges the reasoning paths by reusing their KV-caches to generate the final answer.}
  % \vspace{-8mm}
  \label{fig:main}
\end{figure*}

% As shown in Figure \ref{fig:main}, \name generates result for an input question with two stages. In the parallel reasoning stage, \name generates multple reasoning processes for the question separately. After that, \name generates the final answer by analyzing these diverse reasoning paths based on the intermediate computing results (KV-cache) without re-prefilling
As shown in Figure~\ref{fig:main}, \name generates an answer for a given question in two stages. In the parallel reasoning stage, \name generates multiple independent reasoning trajectories. In the summarization stage, it analyzes these diverse reasoning paths and then fuses them into a final answer. \name utilizes intermediate KV-cache representations from the reasoning stage, which eliminates the need for costly re-prefilling during summarization.

\subsection{Preliminaries}
% w = \{w_i\}_{i=1}^l$, the corresponding embeddings are $
We denote an LLM by $\pi_{\theta}$, where $\theta$ is the set of model parameters. Given an input prompt of $l$ tokens $x = \{x_i\}_{i=1}^l$. The LLM then autoregressively generates an output sequence $y = (y_1, y_2, \dots, y_L)$ with the conditional probability:
\begin{equation}
\pi_{\theta}(y|x) = \prod_{t=1}^{L} \pi_{\theta}(y_t|x, y_{<t})
\end{equation}
For tasks requiring multi-step reasoning, the output $y$ can be decomposed into a reasoning path $r$ followed by a final answer $a$: $y = (r, a)$. During decoding, each new token $y_t$ requires attention over the full context $x, y_{<t}$, which involves computing Key ($K$) and Value ($V$) tensors. To avoid recomputation when generating each $y_t$, LLMs often use a KV-cache to store $K$/$V$ tensors. 
% Let $k_i^j, v_i^j$ denote the cached key and value for token $i$ at attention layer $j$.

% For complex tasks that require deliberation, this output $y$ is often structured to first present an intermediate reasoning path $r$ (i.e., a chain-of-thought) and then conclude with a final answer $a$. We denote this ordered structure as $y = (r, a)$.

% During the autoregressive decoding phase, generating each new token $y_t$ requires an attention calculation over all preceding tokens in the context $(x, y_{<t})$). Recomputing the Key ($K$) and Value ($V$) tensors for this entire context at every step is computationally expensive. 
% Thus, LLMs employ a \texttt{KV cache}. to store the Key ($K$) and Value ($V$) vectors of each token at each attention layer. This prevents the costly recomputation of these vectors for all preceding tokens when generating each new token. We denote the cached key and value for the $i$-th token at the $j$-th attention layer as $k_i^j$ and $v_i^j$.
\subsection{\name Workflow}

As shown in Figure \ref{fig:main}, our approach extends the sequential reasoning LLM paradigm by first generating a set of $P$ distinct reasoning paths $\{r^{(1)}, r^{(2)}, \dots, r^{(P)}\}$ for a single input $x$ in parallel. Each individual reasoning path $r^{(i)}$ is a sequence of tokens representing a unique line of thought, sampled from the distribution:
\begin{align}
\pi_{\theta}(r^{(i)}|x) = \prod_{t=1}^{L_i} \pi_{\theta}(r^{(i)}_t|x, s^{(i)}, r^{(i)}_{<t}) 
\end{align}
Here, $s^{(i)}$ is a special control token that helps initiate a distinct reasoning path, which will be detailed in Section \ref{sec:special_tokens}. 
After generating these parallel paths, the model synthesizes them to produce a final answer, $a$. This answer is conditioned on both the original prompt $x$ and the complete context of all preceding reasoning paths. Let $\mathcal{R}=(r^{(1)}, r^{(2)}, \dots, r^{(P)})$ be the concatenation of all generated reasoning paths. The final answer $a$ is then sampled from the model as follows:
\begin{align}
    \pi_{\theta}(a|x) = \prod_{t=1}^{La} \pi_{\theta}(a_t | x, \mathcal{R}, a_{<t})
\end{align}
Crucially, ParaThinker leverages the KV-caches from the parallel reasoning stage, eliminating the need to re-prefill the context and thereby offering significant computational savings compared to other methods. 

\subsection{Special Tokens for Boosting Thought Diversity}
\label{sec:special_tokens}

\name needs to ensure diverse reasoning paths to avoid the trap of relying on a single sampled sequence.
To achieve this, we introduce a set of trainable special control tokens: \texttt{<think i>}, \texttt{</think i>}, \texttt{<summary>}, and \texttt{</summary>} for $i \in \{1, \dots, P\}$ to control the parallelization and merging operations. 
The \texttt{<think i>} token (denoted as $s^{(i)}$ in our equations) is placed at the beginning of each reasoning path, which leads the model to generate a distinct trajectory. Thus, the distribution of each reasoning path $\pi_{\theta}(r^{(i)}|x) = \prod_{t=1}^{L_i} \pi_{\theta}(r^{(i)}_t|x, s^{(i)}, r^{(i)}_{<t})$ will be conditioned by $s^{(i)}$. 
The closing \texttt{</think i>} token marks the end of a specific path, and the generation of the final answer is then wrapped within \texttt{<summary>} and \texttt{</summary>} tokens. 
This structured use of control tokens is a simple yet powerful mechanism to guide the model's generation process towards diverse and parallel lines of thought. 
% At the beginning of each reasoning path, \texttt{<think i>}, denoted as $\{s^{i}\}_{i=1}^{P}$, are placed to indicate the beginning of each reasoning path. 
% At the end of each sequence, special token \texttt{</think i> \space (i=1,2,\dots,P)} is also appended, indicating the termination of each reasoning path, moving the generation process into summarizing all reasoning paths and deriving the result, wrapped by \texttt{<summary>, </summary>}. 

\subsection{Thought-Specific Positional Embedding}
\label{sec:embedding}

Merging multiple reasoning paths poses challenges due to positional ambiguity. LLMs distinguish tokens based on their content and positional encoding. When multiple reasoning paths are generated in parallel, tokens at the same relative position (\eg the $t$-th token in each $r^{(i)}$) share identical positional encodings. This causes confusion during summarization, as the model cannot differentiate which reasoning stream a token originated from.

\textbf{Flattened Encoding: }
A naive solution assigns unique absolute positions across all paths: $m = l_x + i \cdot l_{\max} + t$, where $l_x$ is the input length, $i$ indexes the reasoning path, and $t$ indexes the token position within that path. While this resolves positional collisions, it results in large positional indices as $P$ increases. Typical positional encoding mechanisms such as Rotary Position Embedding (RoPE) \citep{su2024roformer} encodes relative positions via rotations, and large index differences $|m - n|$ cause attention scores to decay. As a result, tokens from earlier paths (\ie lower $i$ of $r^{(i)}$) contribute less when generating the final answer, introducing imbalance across paths.
% A naive way to solve this problem is to add a unique positional encoding of $PE_m$ into each $r_t^{(i)}$, such as $m=l_x+i*l_{max}+t$, where $l_{max}$ is the maximum token number for each reasoning path. Consequently, the positions of final answers are $\{l_x+P*l_{max}+t\}_{t=1}^{L_a}$, which will be large numbers when $P$ grows large. %However, this positional encoding can increase $l_{max}\times$ as the number of reasoning path grows. 
% However, widely adopted positional encoding methods like rotary positional encoding often possess the long-term decay, meaning that the value of $(RoPE(q_m))(RoPE(k_n))^T$ decays with the relative distance $m-n$ \citep{su2024roformer}. 
% And $|n-m|$ can range from $(P-1)l_{max}+t$ to $t$ for the distance between the final answer and the tokens within the first reasoning path to the last reasoning path, which makes each reasoning path unequally contribute to the final answer. 
% And when $P$ grows larger, this inequality in reasoning path will grow larger, thus hurting the performance of parallel reasoning LLMs. 
% Thus, resulting in the summary smaller attention score on the former reasoning paths when generating final answer, 

\textbf{Sequence-Aware Positional Embedding: }
% Motivated by learnable positional embedding \citep{vit, devlin-etal-2019-bert}, we 
To address positional ambiguity in multi-response generation tasks, \name separates different reasoning paths by augmenting the RoPE mechanism with learnable thought embeddings $\{T^{(j)}\}_{j=0}^P$. Specifically, we add the $T^{(j)}$ to the key and value embeddings of all tokens within the $i$-th reasoning path, which distinguishes each reasoning path at the summarizing phase. 
% And we add $T^{(0)}$ to the final answer within summarizing phase to keep uniform. 
The thought embedding is added to the key before the RoPE rotation is applied. 
Let $\tilde{k}_t^{(j)} , \tilde{v}_t^{(j)} $ denote the cached key and value for token $t$ at path $j$, respectively, from which the key and value vectors are formed as: 
\begin{align}
\tilde{k}_t^{(j)} &= R_t(k_t^{(j)} + T^{(j)}) \\
\tilde{v}_t^{(j)} &= v_t^{(j)} + T^{(j)}
\end{align}
Here, $l_{max}$ denotes the maximum token number for each reasoning path, and $R_t$ is the corresponding RoPE rotation matrix. 
The dot product attention score between a query $q_n$ from the summary (at local position $n$) and a key $\tilde{k}_t^{(j)}$ from path $j$ (at position $m$) is: 
% The attention score between a query at position $n$ (where $l_x+l_{max}\le n \le l_x+l_{max}+l_a$) from the summarizing phase and a key at position $m$ (from sequence $j$, thus $l_x \le m \le l_x+l_{max}$) is then:

% $$\text{score}(n, m) = (R_n q_n)^T \tilde{k}_m^{(j)} = (R_n q_n)^T [R_m (k_m^{(j)} + T^{(j)})]$$
\begin{align}
    \text{score}(n, m) = (R_n q_n^{(i)})^T \tilde{k}_m^{(j)} = (R_n q_n^{(i)})^T [R_m (k_m^{(j)} + T^{(j)})]
\label{score}
\end{align}

Using the RoPE property $(R_n)^T R_m = R_{m-n}$, Eq. \ref{score} can be simplified into two distinct components:

% $$\text{score}(n, m) = \underbrace{q_n^{(i)T} R_{m-n} k_m^{(j)}}_{\text{Content-to-Content}} + \underbrace{q_n^{(i)T} R_{m-n} T^{(j)}}_{\text{Content-to-Segment}}$$
\begin{align}
    \text{score}(n, m) = \underbrace{q_n^{T} R_{m-n} k_m^{(j)}}_{\text{Content-to-Content}} + \underbrace{q_n^{T} R_{m-n} T^{(j)}}_{\text{Content-to-Segment}}
\end{align}
The Content-to-Content term is the standard RoPE attention score, which calculates the relevance between the query's content ($q_n$) and the key's content ($k_m^{(j)}$). This term is not related to the reasoning path number $j$ and thus does not change when scaling parallel reasoning paths. % Pote that $n-m\le l_{max}+l_a$, then $q_n^{T} R_{m-n} k_m^{(j)}$ doesn't decay with the number of reasoning paths, resulting in an acceptable valued decay. %modulated by the rotation matrix $R_{m-n}$ which encodes their relative distance. This component allows the model to understand semantic relationships in the context of their positions. 
Content-to-Segment term calculates the relevance between the query's content ($q_n$) and the learnable identity of the key's entire reasoning path ($T^{(j)}$). This allows the query to directly probe for the origin of the information. 
Because each reasoning path has a unique, learned thought embedding, this term provides an unambiguous signal for the model to differentiate between parallel streams of text, solving the positional ambiguity.

\section{Training and Deployment of \name}
% We adopt a training pipeline from supervised finetuning (SFT) to reinforcement learning (RL), similar to prior works \citep{deepseekr1}, where SFT enables LLMs to understand how to think in parallel, and RL inspires its potential to reasoning. The special challenges include: 1. How to let LLMs know the meanings of \texttt{<think i>} tokens and generate diverse reasoning paths during training? 2. How to enable reinforcement learning for parallel thinking? We will discuss the solutions for these challenges in Section \ref{sec:sft} and Section \ref{sec:rl}. 

We use SFT to train \name. The special challenges include: 
(1) How to let LLMs know the meanings of \texttt{<think i>} tokens and generate diverse reasoning paths during training? (Section \ref{sec:data})
(2) How to design attention mask for \name training? (Section \ref{sec:mask})
(3) How to design inference engine of \name? (Section \ref{sec:infra})

\subsection{Scalable Train Data Curation}
\label{sec:data}
% \label{sec:sft}
To effectively enable native parallel reasoning, we design a scalable training data curation pipeline that consists of two key components: multi-path training data scaling and extensible special tokens training.

\textbf{Multi-Path Training Data Scaling: }
We develop a simple yet effective high-quality parallel reasoning dataset by sampling multiple times from teacher reasoning LLM (\eg DeepSeek-R1~\citep{deepseekr1}). For a query $x$ and groundtruth answer $a$, let the $\hat P$ sampled answers denoted as 
$\{(r^{(1)}, a^{(1)}), (r^{(2)}, a^{(2)}), \dots,$ 
$(r^{(\hat P)}, a^{(\hat P)})\}$. 
We then concat each parallel answer into a groundtruth answer with the format of: 
% $\hat y$ = (\text{<think 1>}$r^{(1)}$\text{</think 1>}
% \text{<think 2>}$r^{(2)}$\text{</think 2>, ...,}
% \text{<think $\hat P$>$r^{(\hat P)}$</think $\hat P$>}
% \text{<summary>}$a$\text{</summary>}). 
$\hat y$ = (\texttt{<think 1>}$r^{(1)}$\text{</think 1>}
\text{, ...,}
\texttt{<think $\hat P$>$r^{(\hat P)}$</think $\hat P$>}
\texttt{<summary>}$a$\texttt{</summary>}). 
The resulting $Data_{r=1}^{\hat P}=(x, \hat y)$ pairs are then used for SFT. 

\textbf{Extensible Special Tokens Training: }
Due to the high cost of teacher LLM inference, we are often faced with the situation where we cannot generate enough reasoning paths when creating $Data_{sft}$, that is, $\hat P$ is less than $P$ in many cases. Thus, during SFT stage, LLMs have to learn to extend to $(r^{(\hat P+1)}, \dots, r^{(P)})$ with training data $Data_{r=1}^{\hat P}$. 
Therefore, we develop a dynamic special token sampling method for extensible special tokens training. Specifically, at each training batch, we randomly sample $\hat P$ special tokens from \texttt{<think~i>}$, i\in\{1, \dots,P\}$. These sample $\hat P$ tokens are placed at the beginning of each reasoning sequence, and the special tokens embeddings are trainable. Thus the special token embeddings are diverse after enough training, and lead LLMs to different reasoning directions. % LLMs can learn that all special tokens \texttt{\textless thinki>}$, i\in\{1, \dots,P\}$ are equal for embracing diverse thoughts, and that 

\subsection{Attention Mask Design}
\label{sec:mask}
% \name adopts a modified attention mask that isolates different reasoning paths and integrates them in the summarization phase. 
To enable effective parallel reasoning, \name adopts a two-phase attention mask design. During the reasoning phase, each reasoning path is decoded independently, with attention restricted to the input prompt and its own token history. Let $M_{i,j}$ denote the attention mask between the index $i$ and index $j$, where attention score can be calculated as: 
$A_{i,j} = \text{Softmax}\left( \frac{q_i \cdot k_j + M_{i,j}}{\sqrt{d_k}} \right)$. 
The attention mask for the $i$-th reasoning path ($r^{(i)}$) is defined as:
\begin{align}
    M^{r^{(i)}}_{t, j} =
\begin{cases}
0, & \text{if } j \leq t \text{ and } j \in \{1, \ldots, l_x\} \cup \text{Ind}_i \\
- \infty, & \text{otherwise}
\end{cases}
\end{align}
where $l_x$ is the length of the input prompt and $\text{Ind}_i$ is the index range for tokens in the $i$-th reasoning path. This enforces independence across reasoning paths by blocking inter-path attention.

During the summarization phase, where each answer token attends to the entire prompt, all reasoning paths, and previously generated answer tokens. The summarization attention mask is defined as:

\begin{align}
M^{A}_{t, j} =
\begin{cases}
0, & \text{if } j \leq t \text{ and } j \in \{1, \ldots, l_x\} \cup \bigcup_{i=1}^P \text{Ind}_i \cup \text{Ind}_a \\
- \infty, & \text{otherwise}
\end{cases}
\end{align}
where $\text{Ind}_a$ denotes the index range of the answer tokens. This mask allows the final answer to integrate all parallel thoughts without violating autoregressive constraints.

\subsection{Inference Engine}
\label{sec:infra}
The inference engine for \name is built upon the vLLM framework \citep{vllm} to leverage its efficient parallel scaling. % , particularly PagedAttention, for parallel decoding. 
The inference process is divided into two distinct phases: 

\textbf{Parallel Reasoning Phase: }The engine processes the $P$ reasoning paths concurrently as a single batch. This synchronous decoding phase terminates for all paths simultaneously as soon as the one of the following conditions is met:
(1) Path Completion: Any one of the $P$ paths generates an end-of-sequence (EOS) token. 
(2) Budget Exceeded: Any one of the $P$ paths reaches the predefined token budget, B. 
This uniform termination strategy ensures all reasoning paths maintain an equal length, preventing processing imbalance. As empirically justified in Section \ref{sec:termination}, this approach yields the highest accuracy.

\textbf{Summarization Phase: }Following the parallel reasoning phase, the engine constructs an attention context spanning the KV caches of all $P$ reasoning paths, eliminating the need for costly re-prefilling. 
Leveraging vLLM's PagedAttention, this step is performed with zero data copying, as the summary sequence can directly reference the memory blocks of all preceding paths. 
The final answer is then autoregressively decoded from this unified context.

%% file: tex/experiment.tex
\section{Experiment}
\label{sec:exp}
In this section, we conduct experiments to address the following research questions:
\begin{itemize}[leftmargin=*, itemsep=0pt, parsep=0pt, topsep=2pt]  % This option removes the indentation
    \item \textbf{Scalability and Performance: }How does \name's reasoning accuracy scale with an increase in parallel reasoning paths number and generation budget? (Section \ref{sec:scalability})
    % \item  \textbf{Is \name applicable to different training techniques (RL/SFT-trained LLMs) or different scaling methods (majority voting)? \ref{sec:other})}
    \item \textbf{Inference Efficiency: }What are the trade-offs in terms of inference latency and memory consumption as we scale the number of parallel paths? (Section \ref{sec:efficiency})
    \item \textbf{Fine-Grained Analysis: }What is the optimal termination strategy for parallel reasoning stage, and how much does each component contribute? (Section \ref{sec:termination} and Section \ref{sec:ablation})
    % \item \textbf{RQ2:} What is the relationship between sequential test-time scaling and native parallel scaling? Can \name improve performance by scaling the parallel size independently of sequential test-time scaling? Furthermore, how \name organically combine sequential and parallel scaling to leverage the model's capabilities effectively under a fixed token budget?
    % \item \textbf{RQ3:} Does the theoretically proven sequence-aware positional embeddings have a great effect in practice?
\end{itemize}

\subsection{Experimental Setup}
\label{sec:exp_setup}
\textbf{Training Details: }Our experiments are based on a Qwen-2.5 \citep{qwen2025qwen25technicalreport} 1.5B and 7B parameter model distilled from DeepSeek-R1 \citep{deepseekr1}, which we denote as original models below. 

For SFT, we construct a parallel reasoning dataset with 6.2K problem-solution pairs. 3.5K of the problems are sampled from the Open-R1 \citep{openr1} filtered to include only those with more than 4 existing answer variations. We also randomly sample 1.5K from and DeepMath \citep{he2025deepmath} dataset, which provides 3 answers per question, and 1.2K from s1k \citep{muennighoff2025s1} (0.4K filtered for clear answers) and limo \citep{ye2025limo} (0.8K full dataset). To enrich diversity, we use gpt-oss-20b \citep{openai2025gptoss120bgptoss20bmodel} as a teacher model, generating additional solutions at temperature 0.8, yielding six reasoning paths per problem. Each instance thus consists of a query ($x_i$), ground-truth answer ($a_i$), and $\hat P=6$ distinct reasoning paths.
%To avoid the influence of teacher model and new training data, we implement an ablation experiment to show the effectiveness of our method in Section \ref{sec:ablation}. 
%The final training data is a set of tuples $\tilde D=\{x_i, (r^{(1)}_i, r^{(2)}_i, \dots, r^{(\hat P)}_i, a_i)\}_{i=1}^{|D|}$, where each instance contains the problem query ($x_i$), the ground-truth answer ($a_i$), and $\hat P=6$ distinct reasoning paths that include both the original solutions and the newly generated ones. 
% Finally, each instance in the dataset contains the problem query ($x_i$), the ground-truth answer ($a_i$), and $\hat P=6$ distinct reasoning paths that include both the original solutions and the newly generated ones.
% The prompt structure used for generation is detailed in Appendix \ref{sec:prompt}.

During fine-tuning, we use a maximum context length of 28K tokens. The model is trained for 2-3 epochs on multiple A800 GPUs. More training details can be found in Appendix \ref{appendix:train_settings}. 
During every step, we randomly choose a path number $P$ from the set $\{2, 4, 6\}$ and construct a training sample by concatenating $P$ samples, which is detailed in Appendix \ref{appendix:sft_data}.

For data ablation, we ensure comparable dataset volume by unrolling our parallel dataset into $\tilde D_{ablation}=\{\{x_i,(r^{(j)}_i,a_i)\}_{j=1}^{\hat P}\}_{i=1}^{|D|}$, then fine-tune the 1.5B baseline with identical settings. We then report results for \textit{sequential}, \textit{majority voting}, and \textit{re-prefilling} against \name-1.5B.

\textbf{Baselines: }
% For a fair comparison, baselines are built upon the same base model $\pi_\theta$, which undergoes an equivalent training procedure. To ensure the training data is comparable in volume, the SFT dataset for the baseline is constructed by unrolling our parallel dataset: 
% $\tilde D_{baseline}=\{\{x_i, (r^{(j)}_i, a_i)\}_{j=1} ^{\hat P} \}_{i=1}^{|D|}$. 
% After training on $\tilde D_{baseline}$, we denote the LLM as $\tilde \pi_\theta$. We compare \name against two primary baselines operating on $\tilde \pi_\theta$: 
We compare \name against: 
\begin{itemize}[leftmargin=*, itemsep=0pt, parsep=0pt, topsep=2pt]  % This option removes the indentation
    \item \textit{Sequential: }Direct reasoning with original 1.5B/7B models.
    \item \textit{Majority Voting: }Generate $P$ independent paths and return the majority answer \citep{majority_voting}. 
    \item \textit{Re-Prefilling: }Generate $P$ paths, concatenate them, and feed the full context into the model for summarization. This mimics \name’s summarization but is inefficient since KV caches are not reused (Section \ref{sec:efficiency}). % This baseline also generates $P$ paths, concatenates them, and then feeds the entire context back into the model to produce a final summary. This approach is functionally analogous to \name's summarization stage but is computationally inefficient, as it does not reuse the KV cache. (Section \ref{sec:efficiency})
\end{itemize}

% After training on $\tilde D_{baseline}$, we denote the LLM as $\tilde \pi_\theta$. 

%\textbf{Evaluation Setup: } 
% Our system is implemented using the vLLM inference framework \citep{vllm}, integrated with our custom parallel generation engine. We use a sampling temperature of \xx with a top-p value of \xx. To account for output randomness, we report pass@1 accuracy, calculated as $pass@1=\frac{1}{k}\sum_{i=0}^k p_i$, where $p_i$ s a binary indicator of correctness for the $i$-th response. Following DeepSeek-R1, we set $k$ depending on the test set size, thus we set $k=16$ to AIME 2024, AIME 2025, AMC 2023, and $k=4$ to MATH-500. \syf{I have modified the setting of k here}
% Because the randomness of LLM outputs, we calculate {pass@1} as $pass@1=\frac{1}{k}\sum_{i=0}^k p_i$, where $p_i$ denotes the correctness of the $i$-th response. We set $k$ as \xx in our experiments. 
% We use a sampling temperature of \xx and a top-$p$ value of \xx. 
% For majority voting \citep{majority_voting} of $n$ results, we denote them as {maj@$n$}. 

\textbf{Benchmarks and Evaluation Setup: }
We evaluate our model on four challenging mathematical reasoning benchmarks: AIME 2024, AIME 2025, AMC 2023, and MATH-500 \citep{math500}. We employ two primary criteria for evaluation: effectiveness, measured by accuracy; and efficiency, measured by performance under a fixed token budget. To assess efficiency, we use a budget control method where each reasoning path is limited to a maximum of $B$ tokens ($|r^{(i)}|\le B$). 
If a model reaches this budget without naturally stopping, we enforce termination and then initiate the summarization stage by adding the (\texttt{<summary>}) token. This allows us to examine the utilization of the test-time scaling budget.% Efficiency is measured by performance under a fixed token budget, $B$, which is the maximum number of reasoning tokens allowed per path $|r^{(i)}|\le B$. 
% We employ two criterions of effectiveness and efficiency, where effectiveness is measured by the accuracy of LLMs, and efficiency is the performance of LLMs under a token budget. % We implement the method of \xx in our method and baselines to deal with token budget. 
% Specifically, we conducted a budget control method to examine whether LLMs have achieved better budget utilization for testing time scaling. Each budget $B$ is the maximum token number of reasoning for LLMs. After the LLM reaches the token number budget and doesn't stop naturally, we terminate the thinking stage by adding the end-of-thinking delimiter (\texttt{</think>}) and begin the summary by adding (\texttt{<summary>}). 
% For multiple reasoning paths, without otherwise stated, the budget is set for each reasoning path, namely $|r^{(i)}|\le B$. 

Our system is implemented using the vLLM inference framework \citep{vllm}, integrated with our custom parallel generation engine. For the 1.5B parameter model, we employ a sampling temperature of 0.5 and a top-p value of 1.0, while for the 7B parameter model, we use a temperature of 0.6 and a top-p value of 1.0. To account for output randomness, we report pass@1 accuracy, calculated as $pass@1=\frac{1}{k}\sum_{i=0}^k p_i$, where $p_i$ is a binary indicator of correctness for the $i$-th response. Following DeepSeek-R1, we set $k$ depending on the size of test dataset, thus we set $k=16$ for AIME 2024, AIME 2025, AMC 2023, and $k=4$ for MATH-500.

% \textbf{Length Extrapolate. }

\begin{table}[t!]
\centering
\small
\label{tab:main}
\begin{tabular*}{\textwidth}{@{\extracolsep{\fill}}lccccc}
\toprule
& AIME 2024 & AIME 2025 & AMC 2023 & MATH-500 & Average \\
\midrule
\rowcolor{gray!20}
\multicolumn{6}{l}{\textit{Original Model: DeepSeek-R1-distill-Qwen-1.5B}}\\
\midrule
Sequential (16K) & 26.1 & 22.4 & 67.1 & 81.2 & 49.2 \\
Sequential (32K) & 28.3 & 24.5 & 68.9 & 81.8 & 50.9 \\
Sequential (64K) & 27.1 & 25.5 & 67.7 & 81.7 & 50.5 \\
Sequential (128K) & 27.4 & 22.1 & 68.0 & 81.8 & 49.8 \\ \cmidrule{1-6}
Majority (2$\times$16K) & 25.9 & 23.0 & 67.0 & 81.4 & 49.3 \\
Majority (4$\times$16K) & 32.9 & 27.5 & 74.3 & 86.7 & 55.4 \\
Majority (8$\times$16K) & 41.0 & 31.8 & 79.8 & 89.0 & 60.4 \\ \cmidrule{1-6}
Reprefill (2$\times$16K) & 30.4 & 26.7 & 70.6 & 60.8 & 47.1 \\
Reprefill (4$\times$16K) & 24.2 & 25.8 & 61.3 & 58.6 & 42.5 \\
Reprefill (8$\times$16K) & 14.2 & 13.3 & 60.0 & 55.3 & 35.7 \\ \cmidrule{1-6}
\textbf{\name-1.5B} (2$\times$16K) & 34.8 & 24.2 & 73.1 & 87.5 & 54.9 \\
\textbf{\name-1.5B} (4$\times$16K) & 43.3 & 26.7 & 80.8 & 88.7 & 59.9 \\
\textbf{\name-1.5B} (8$\times$16K) & 48.1 & 31.9 & 83.1 & 89.7 & 63.2 \\
\midrule
\rowcolor{gray!20}
\multicolumn{6}{l}{\textit{Original Model: DeepSeek-R1-distill-Qwen-7B}}\\
\midrule
Sequential (16K) & 51.9 & 37.9 & 88.4 & 91.2 & 67.4 \\
Sequential (32K) & 55.5 & 37.9 & 89.8 & 92.0 & 68.8 \\
Sequential (64K) & 56.0 & 39.6 & 89.8 & 92.5 & 69.5 \\
Sequential (128K) & 52.7 & 40.4 & 89.8 & 92.6 & 68.9 \\ \cmidrule{1-6}
Majority (2$\times$16K) & 52.3 & 38.3 & 88.4 & 91.4 & 67.6 \\
Majority (4$\times$16K) & 60.6 & 43.1 & 92.2 & 93.5 & 72.4 \\
Majority (8$\times$16K) & 68.8 & 49.6 & 93.1 & 94.2 & 76.4 \\ \cmidrule{1-6}
Reprefill (2$\times$16K) & 42.9 & 33.8 & 88.1 & 63.8 & 57.2 \\
Reprefill (4$\times$16K) & 43.3 & 33.3 & 86.3 & 63.2 & 56.5 \\
Reprefill (8$\times$16K) & 43.3 & 31.7 & 91.9 & 63.7 & 57.7 \\ \cmidrule{1-6}
\textbf{\name-7B} (2$\times$16K) & 57.1 & 46.0 & 89.5 & 93.2 & 71.5 \\
\textbf{\name-7B} (4$\times$16K) & 63.3 & 46.9 & 91.7 & 94.2 & 74.0 \\
\textbf{\name-7B} (8$\times$16K) & 68.8 & 51.3 & 93.3 & 94.5 & 77.0 \\
\midrule
\end{tabular*}
\caption{Accuracy of \name and baselines on MATH-500, AIME 2024, AIME 2025, and AMC 2023. We report Pass@1 accuracy (\%). Values in brackets indicate the maximum generation length $L$ (\eg 16K); for parallel generation methods, we use $P \times L$ to denote generating $P$ reasoning paths, each with a maximum length of $L$.}
\label{tab:main}
\end{table}

\subsection{Scalability and Performance}
\label{sec:scalability}

% Table \ref{tab:main} compares \name with baselines under different total token budgets. 
% Compare to the original sequential LLM, \name improves the accuracy of 1.5B and 7B LLMs for 14.5\% and 8.3\% on AIME 2024, 3.6\% and 8.8\% on AIME 2025 at each same budget compared to the original LLM (\eg $2\times 16K$ vs. $32K$), showing the effectiveness of parallel reasoning. 
% \name outperforms both majority voting and re-prefilling in most cases as the number of paths increases. 
% In average, accuracy of \name is 4.3\% and 2.0\% higher in accuracy compared to majority voting on 1.5B and 7B LLMs respectively. 
% Besides, for $P=2$, \name significantly outperforms maj@2, indicating that it does not merely rely on selecting the majority outcome but instead learns to synthesize information across paths. 
% This suggests that its summarization mechanism learns a more sophisticated aggregation strategy than simply counting votes.
% The performance of the re-prefilling baseline degrades with more paths, which we attribute to the context length limitations of the flattened positional encoding scheme discussed in Section \ref{sec:embedding}. 
Table \ref{tab:main} compares \name with baseline methods under different token budgets. 
Compared with sequential LLMs, \name improves accuracy by up to 14.5\% (1.5B) and 8.3\% (7B) on AIME 2024, and by 3.6\% (1.5B) and 8.8\% (7B) on AIME 2025 at each budget on average (\eg $2\times16$K vs.\ 32K), demonstrating the effectiveness of parallel reasoning. 
On average, \name achieves 4.3\% and 2.0\% higher accuracy than majority voting on the 1.5B and 7B models, respectively. 
This indicates that the summarization stage captures a richer aggregation strategy than vote counting. 
Besides, for $P=2$, \name substantially outperforms maj@2 (which randomly selects one result from two reasoning paths and thus closely resembles sequential reasoning), showing that it does more than simply pick the majority answer and instead learns to integrate information across reasoning paths. 
By contrast, the performance of the re-prefilling baseline degrades with more paths, which we attribute to the context length limitations of the flattened positional encoding scheme discussed in Section \ref{sec:embedding}. 

We analyze how performance scales with the number of parallel paths in Table \ref{tab:scale}, where increasing the path count consistently yields higher accuracy at larger generation budget. For sequential reasoning LLM ($P=1$), expanding the token budget beyond 32K yields no further accuracy gains, whereas \name continues to improve. These results indicate that \name effectively extends the scaling law beyond the point where sequential reasoning models typically encounter a test-time scaling bottleneck. 

\begin{table}[t!]
\centering
\small
\begin{tabular}{c|ccccc}
\toprule
& $B=8K$ & $B=16K$ & $B=32K$ & $B=64K$ & $B=128K$ \\
\midrule
$P=1$ & \cellcolor{blue!36}23.5 & \cellcolor{blue!43}26.1 & \cellcolor{blue!48}28.3 & \cellcolor{blue!45}27.1 & \cellcolor{blue!46}27.4 \\
$P=2$ & \cellcolor{blue!47}27.1 & \cellcolor{blue!53}29.2 & \cellcolor{blue!67}34.8 & \cellcolor{blue!69}35.8 & \cellcolor{blue!42}25.0 \\
$P=4$ & \cellcolor{blue!25}18.1 & \cellcolor{blue!55}30.2 & \cellcolor{blue!75}38.1 & \cellcolor{blue!88}43.3 & \cellcolor{blue!71}36.7 \\
$P=8$ & \cellcolor{blue!0}7.9 & \cellcolor{blue!35}22.3 & \cellcolor{blue!67}35.0 & \cellcolor{blue!83}41.5 & \cellcolor{blue!90}48.1 \\
\bottomrule
\end{tabular}
\caption{Accuracy of baseline ($P=1$) and \name-1.5B ($P=2/4/8$) on AIME 2024 with different reasoning path number \textit{P} and total token budget \textit{B} (total token length of all reasoning paths). }
\label{tab:scale}
\end{table}

We further analyze the relationships between \name and majority voting in detail. 
The result is shown in Table \ref{tab:majority_voting}. 
We find that our method does not conflict with the majority voting, \name with majority voting can achieve higher accuracy than only using \name. 
The highest accuracy of \name-1.5B+maj@8 can achieve accuracies of 66.7\% and 60.0\% on AIME 2024 with $P=4$ and $P=8$, gaining 23.4\% and 11.9\% accuracy improvements against pass@1. 
Besides, majority voting is not able to be applied to scenarios where results can't be quantified (\eg coding, document generation, etc.), which we will evaluate \name as a future work.

\begin{table}[t!]
\centering
\small
\begin{tabular}{l|c|c|c|c}
\toprule
\textbf{} & $P$=1 & $P$=2 & $P$=4 & $P$=8 \\
\midrule
\textit{pass@1}& 26.1 & 34.8 & 43.3 & 48.1 \\
\textit{maj@4}& 32.9 & 42.5 & 53.0 & 56.3 \\
\textit{maj@8}& 41.0 & 50.1 & 61.7 & 59.9 \\
\textit{maj@16}& 47.8 & 56.7 & 66.7 & 60.0 \\
\bottomrule
\end{tabular}
\caption{\name-1.5B together with majority voting on AIME 2024. $P$: number of parallel reasoning paths; $maj@k$: majority voting with $k$ samples. }
\label{tab:majority_voting}
\end{table}

% We first compare the performance between the baseline LLM and \name of different reasoning path numbers. 
% As illustrated in Figure \xx, both \name and baseline model achieve higher accuracy with larger reasoning budget. 
% But for \name, there is a third dimension - the number of parallel paths, where LLMs with larger parallel number achieve better performance under each token budget. 

% The accuracy of \name compared with more baselines under \xx budget is shown in Table \xx. 
% We can observe that \name consistently outperforms baselines under different number of reasoning paths (v.s. number of majority voting / re-prefilling paths). 
% This shows that \name learns how to summarize the reasoning paths instead of simply believing in the reasoning results in majority. 
% For re-prefilling method, the accuracy of larger reasoning paths drops because of the long context length problem of flattened encoding stated in Section \ref{sec:embedding}. 

% \subsection{\name+: parallel thinking with other techniques}
% \label{sec:other}
% We further analyze whether \name doesn't conflict with different training techniques or base LLMs. Besides, we want to check out what is the relationship between \name and majority voting. 

% Experiments show that powered by \name, SOTA LLMs or majority voting results can achieve even higher accuracy. 

\subsection{Inference Efficiency}
\label{sec:efficiency}
\begin{figure*}[htbp]
  \centering
  \includegraphics[width=0.6\linewidth]{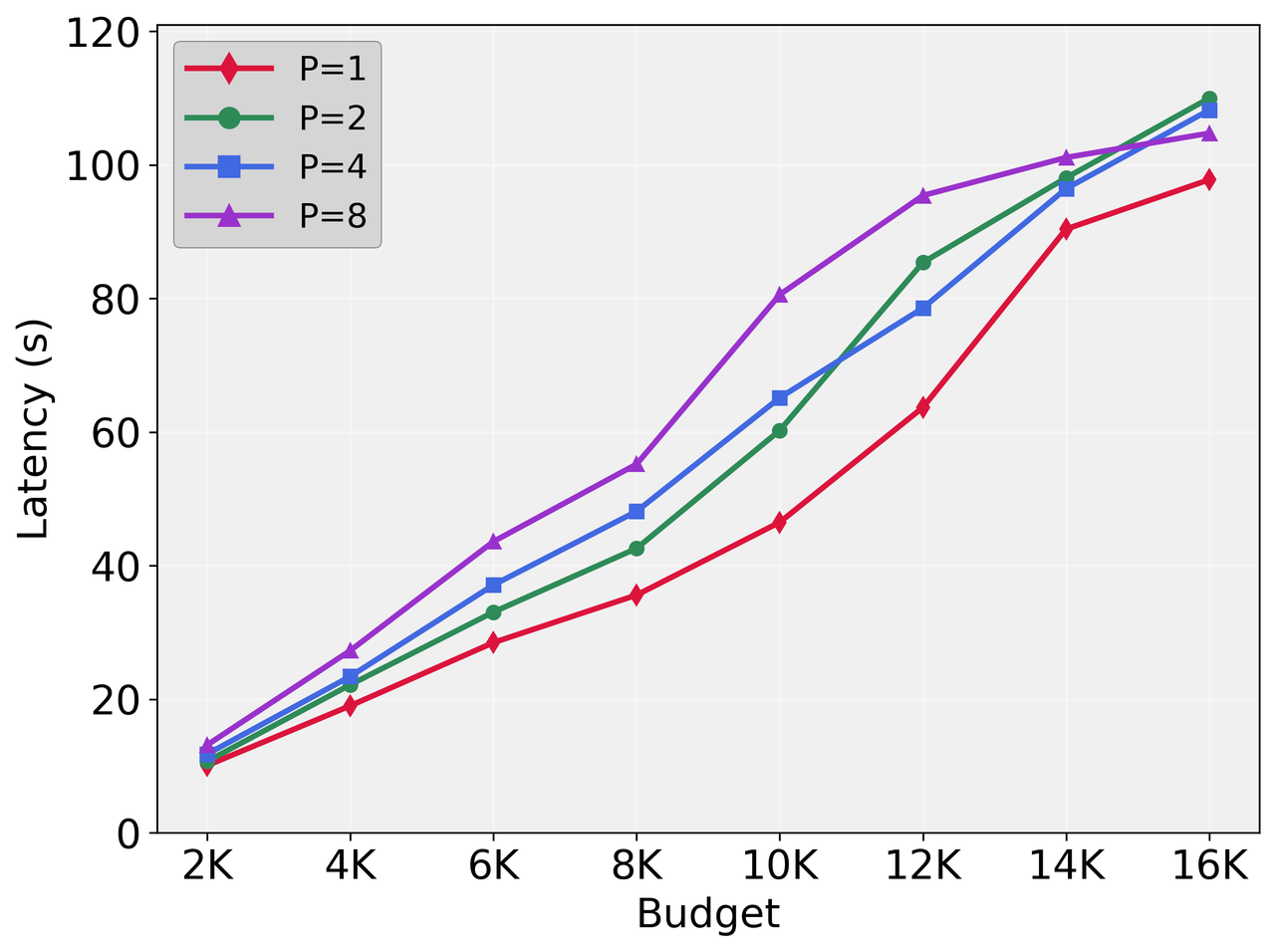}
  % \vspace{-8mm}
  \caption{Total latency of \name-1.5B (batch size=1) with different number of parallel paths ($P$) under different generation budgets for each path (\ie total decoding latency of $P\times B$ tokens for \name with $P$ paths. }
  \label{fig:efficiency}
\end{figure*}

A key advantage of \name is its efficiency. % against pre-profilling and Best-of-N 
% We measure the latency and memory footprint as a function of the number of parallel paths $P$. 
Figure \ref{fig:efficiency} shows that latency does not grow linearly with the number of paths $P$, and with $P$ grows, the total inference latency increases slightly. 
% Figure \xx shows the inference latency of \name against the number of reasoning paths under the batch size 1. 
This is because the decoding phase is typically bounded by memory bandwidth, and increasing the number of parallel reasoning paths does not increase data movement operations. 
The inference latency of \name even decreases slightly under some budgets when number of parallel size grows, because we terminate the reasoning process once the first reasoning path stops, and when the number of paths increases, the probability of \name terminating earlier grows larger. 
The efficiency of our method gives us the proof that we can achieve greater accuracy through parallel scaling within acceptable inference latency. 

% In contrast, the Re-prefilling baseline shows a prohibitive, near-linear growth in latency, rendering it impractical for high values of $P$. 

% Memory consumption for ParaThinker also scales gracefully, confirming its viability for deployment in resource-constrained environments. 
% The efficiency of our method gives us the proof that we can achieve greater accuracy through parallel scaling within acceptable inference latency. 

% Figure \xx illustrates the memory consumption of \name with different reasoning paths. 

% \subsection{Effectiveness analysis}
% \label{sec:analysis}

% \begin{table}[t!]
% \centering
% \small
% \begin{tabular}{l|c|c|c|c}
% \toprule
% \textbf{} & $P$=1 & $P$=2 & $P$=4 & $P$=8 \\
% \midrule
% \textit{pass@1}& 51.9 & 57.1 & 63.3 & 68.8 \\
% \textit{maj@4}& 60.6 & 64.9 & 70.3 & 74.7 \\
% \textit{maj@8}& 68.8 & 68.9 & 72.1 & 77.6 \\
% \textit{maj@16}& 73.3 & 70.0 & 73.3 & 76.7 \\
% \bottomrule
% \end{tabular}
% \caption{\name-7B together with majority voting on AIME 2024. $P$: number of parallel reasoning paths; $maj@k$: majority voting with $k$ samples. }
% \label{tab:majority_voting}
% \end{table}

\subsection{Termination Strategies for the Parallel Reasoning Stage}
\label{sec:termination}

\begin{table}[t!]
\centering
\small
\begin{tabular}{l|c|c|c}
\toprule
\textbf{} & $P$=2 & $P$=4 & $P$=8 \\
\midrule
Last-Finish & 32.1 & 37.1 & 42.5 \\
Half-Finish & 34.8 & 38.3 & 42.5 \\
First-Finish (Default) & 34.8 & 43.3 & 48.1 \\
\bottomrule
\end{tabular}
\caption{Accuracy of \name-1.5B on AIME 2024 under budget $B$ for each reasoning path based on different strategies for terminating the parallel reasoning stage before proceeding to summarization. }
\label{tab:termination}
\end{table}

We compare three strategies for terminating the parallel reasoning stage before proceeding to summarization:  
(1) \textit{Last-Finish:} Wait for all $P$ paths to complete. 
(2) \textit{Half-Finish:} Terminate when $P/2$ paths have completed. 
(3) \textit{First-Finish:} Terminate when the first path completes (our default strategy).

% We set the temperature of 0.5 for \name. 
As shown in Table \ref{tab:termination}, the First-Finish strategy yields the best performance. We attribute this to the fact that it maintains equal reasoning lengths across all paths, preventing any single path from dominating the context and ensuring a balanced contribution to the summarization stage. It is also, by definition, the most computationally efficient strategy.

\subsection{Ablation Study}
\label{sec:ablation}

To have a fine-grained analysis of our key design choices, we conduct an ablation study on AIME 2024 based on \name-1.5B. 
% The ablation results are summarized in Table \ref{tab:ablation}. 
% Given the significant computational cost of full RL training, we conduct our component ablations on SFT-only models. We posit that the architectural insights from this controlled setting are fundamental and representative of the trends that would be observed in the fully-trained model. \syf{Add YaRN ablation}
% Due to the high training cost of RL, we only use SFT for the ablation study of special tokens and thought embeddings, where both our method and the ablation methods are only trained using the SFT dataset and method. 
% We believe that the LLM trained using the whole pipeline can show the similar pattern with the SFT method. 

\begin{itemize}[leftmargin=*, itemsep=0pt, parsep=0pt, topsep=2pt]  % This option removes the indentation
    \item \textbf{Train Data: }We test whether the performance gain of \name comes from the training dataset. The training details of data ablation are stated in Section \ref{sec:exp_setup}, where we use all the data from our dataset (6 samples for each question) to finetune the original LLM, remaining all other settings the same. %The results are shown in Table \ref{tab:dataset-ablation}. The results of finetuned LLM are similar to (even slightly worse than) the original LLM. \name outperforms baselines on each token budget, showing the effectiveness of our method. 
    Table \ref{tab:dataset-ablation} indicates that finetuning does not improve performance, with results even slightly worse than the original LLM. \name, on the other hand, outperforms all baselines across budgets, confirming its effectiveness.
    % \item \textbf{Special Tokens: }We remove special tokens mentioned in Section \ref{sec:special_tokens} and rely solely on a high sampling temperature \xx to induce diversity. The resulting performance drops, demonstrating that temperature alone is insufficient. The conditioned generation guided by our special tokens is crucial for producing a comprehensive set of diverse thought paths. 
    % Removing the \texttt{<think i>} tokens leads to a significant drop in performance. This show that only using temperature for thought diversity is not enough. Our design of thought tokens further encourage the diversity by conditioned generation, which results in more comprehensive though paths. 
    \item \textbf{Thought Embedding: }We conduct two ablation studies of thought embedding: (1) removing the embedding entirely and (2) replacing it with a naive flattened positional encoding as stated in Section \ref{sec:embedding}. 
    As shown in Table \ref{tab:ablation}, the flattened encoding leads to a severe accuracy drop, especially with larger budgets, confirming the detrimental effect of long-range positional decay. Surprisingly, removing the embedding altogether, while still underperform our proposed method, is superior to the flattened approach. This suggests the model can partially infer path distinctions from context but is misled by the ambiguous signals of a naive long-range positional encoding. 
\end{itemize}

\begin{table}[t!]
\centering
\small
\label{tab:dataset-ablation}
\begin{tabular*}{\textwidth}{@{\extracolsep{\fill}}lccccc}
\toprule
& AIME 2024 & AIME 2025 & AMC 2023 & MATH-500 & Average \\
\midrule
\rowcolor{gray!20}
\multicolumn{6}{l}{\textit{DeepSeek-R1-distill-Qwen-1.5B-SFT}}\\
% \midrule
% & AIME 2024 & AIME 2025 & AMC 2023 & MATH-500 & Average \\
\midrule
Sequential (16K) & 26.3 & 18.5 & 66.0 & 81.1 & 48.0 \\
Sequential (32K) & 22.9 & 22.1 & 64.1 & 77.6 & 46.7 \\
Sequential (64K) & 25.8 & 17.3 & 62.2 & 77.6 & 45.7 \\
Sequential (128K) & 24.8 & 21.9 & 63.6 & 78.6 & 47.2 \\ \cmidrule{1-6}
Majority (2$\times$16K) & 26.0 & 18.1 & 66.3 & 81.0 & 47.9 \\
Majority (4$\times$16K) & 32.2 & 23.4 & 72.1 & 86.5 & 53.6 \\
Majority (8$\times$16K) & 42.5 & 27.1 & 79.8 & 89.2 & 59.7 \\ \cmidrule{1-6}
Reprefill (2$\times$16K) & 23.3 & 16.3 & 65.6 & 76.8 & 45.5 \\
Reprefill (4$\times$16K) & 15.0 & 11.7 & 55.6 & 70.6 & 38.2 \\
Reprefill (8$\times$16K) & 15.8 & 9.2 & 58.8 & 66.6 & 37.6 \\ \cmidrule{1-6}
\rowcolor{gray!20}
\multicolumn{6}{l}{\textit{\name-1.5B}}\\
\midrule
\textbf{\name-1.5B} (2$\times$16K) & 34.8 & 24.2 & 73.1 & 87.5 & 54.9 \\
\textbf{\name-1.5B} (4$\times$16K) & 43.3 & 26.7 & 80.8 & 88.7 & 59.9 \\
\textbf{\name-1.5B} (8$\times$16K) & 48.1 & 31.9 & 83.1 & 89.7 & 63.2 \\
\midrule
\end{tabular*}
\caption{Accuracy of \name and baselines on MATH-500, AIME 2024, AIME 2025, and AMC 2023. We report Pass\@1 accuracy (\%). Values in brackets indicate the maximum generation length $L$ (\eg 16K); for parallel generation methods, we use $P \times L$ to denote generating $P$ reasoning paths, each with a maximum length of $L$.}
\label{tab:dataset-ablation}
\end{table}

\begin{table}[t!]
\centering
\small
\begin{tabular}{l|c|c|c}
\toprule
\textbf{} & $P$=2 & $P$=4 & $P$=8 \\
\midrule
\textbf{ParaThinker-1.5B} & 34.8 & 43.3 & 48.1 \\
\midrule
Thought Embedding Ablation & 33.3 & 39.0 & 46.7 \\
% \textbf{YaRN Extrapolation Technique Ablation} & 32.3 & 37.5 & 42.1 \\
\bottomrule
\end{tabular}
\caption{Ablation study on the effect of thought embedding (AIME 2024, avg@16, $t = 0.5$, $B = 16$K).}
\label{tab:ablation}
\end{table}

%% file: tex/related.tex
\section{Related Work}

% \lyc{some cite commands  should be changed to citep}

\subsection{Sequential Test-Time Scaling}
Recent advances in test-time scaling seek to improve LLM reasoning by increasing computational depth during decoding, primarily through reinforcement learning (RL) \citep{OpenAI2024o1, deepseekr1, gemini25, yang2025qwen3technicalreport, kimiteam2025kimik15scalingreinforcement, zhao2025d1, dapo} and supervised fine-tuning (SFT) \citep{muennighoff2025s1, ye2025limo}. RL-based methods \citep{OpenAI2024o1, shao2024deepseekmathpushinglimitsmathematical} encourage LLMs to allocate more computation to promising reasoning paths by encouraging self-reflection and iterative trial. Other approaches distill long-form rationales from larger teacher models into smaller student models, enabling deeper internal reasoning through fine-tuned CoT supervision \citep{bespoke_stratos, ye2025limo, xu2025redstardoesscalinglongcot, latent_reasoning}. 
While these methods significantly enhance LLM performance on complex tasks, they often suffer from increased inference latency and compute consumption due to long output sequences \citep{Sun2025TimesUp, zhu2025conciseadaptivethinkinglarge, meta_rft, wen2025budgetthinker}. 
Moreover, excessively long reasoning traces may introduce ``overthinking'' effects such as repetition or hallucination \citep{chen2025think23overthinkingo1like, ghosal2025doesthinkinghelpunderstanding}. 
Besides, recent works have also shown that sequential reasoning LLMs are brittle to reasoning order \citep{chen2024premise_order} or shallow token attacks \citep{preemptive_attacks, qi2025safety}. 
In contrast, our method introduces a new dimension of inference-time scaling—width—by executing multiple reasoning paths in parallel and summarizing them. This approach preserves reasoning efficiency while avoiding long single-path decoding.

\subsection{Search-Based Methods for Parallel Reasoning}
Parallel Decoding LLMs improve reasoning by sampling multiple tokens at each step to accelerate LLM inference and/or improving LLM performance.
% and selecting the most plausible one using various heuristics or learned reward models. 
Early techniques include beam search \citep{beamsearch}, self-consistency \citep{wang2023selfconsistency}, speculative decoding \citep{leviathan2023fast, chen2023accelerating} and majority voting \citep{majority_voting}.  
Recent advancements include Best-of-N \citep{best_of_n}, Tree of Thoughts (ToT) \citep{yao2023tree}, and Monte Carlo Tree Search (MCTS) \citep{snell2024scalingllmtesttimecompute, guan2025rstarmathsmallllmsmaster}. These approaches typically require an external verifier to evaluate and rank candidate completions, increasing computational cost and often relying on domain-specific or manually constructed reward signals. 
% These methods often depend on an external trained verifier to judge the answers, thus increasing computational costs and heavily depending on specialized training data (\ie reward signal). 
Our method departs from these paradigms by generating multiple reasoning trajectories internally and merging them using a lightweight summarization step, without requiring external verifiers or retraining.

% Parallel generation methods aim to increase LLM reasoning capacity by generating in parallel. Early techniques include beam search \citep{beamsearch}, self-consistency \citep{wang2023selfconsistency}, and majority voting \citep{majority_voting}. 
% These methods require no additional training of LLMs. 
% Our method does not conflict with these methods and can be used jointly, and we also compare our method with majority voting in Section \ref{sec:exp}. 
% Recent works focus on searching-based methods powered a reward model or heuristics to select the best answer, including Best-of-N \citep{best_of_n}, beam search \citep{yao2023tree, feng2023alphazero}, and Monte-Carlo Tree Search \citep{snell2024scalingllmtesttimecompute, guan2025rstarmathsmallllmsmaster}. 

\subsection{Natively Parallel Generation Methods}
Another line of work focuses on empowering LLMs to generate multiple tokens at each decoding iteration to accelerate LLMs theoretically. Diffusion-based language models \citep{he-etal-2023-diffusionbert, ye2023diffusionllm, zhao2025d1, wang2025remasking, arriola2025block} sample multiple tokens in parallel during each diffusion step. While these methods can theoretically enable parallel generation, recent theoretical analyses \citep{feng2025theoreticalbenefitlimitationdiffusion} shows that for tasks involving sequential dependencies (\eg reasoning), the number of required diffusion steps can scale linearly with sequence length, undermining their efficiency. 
PARSCALE \citep{chen2025parallelscalinglawlanguage} investigates architectural parallelism by duplicating the input multiple times, applying distinct transformations, and aggregating outputs token-wise. However, this approach still requires architectural changes and specialized continual pretraining. In contrast, our approach retains the standard LLM architecture and introduces parallelism at the reasoning level by generating and caching multiple distinct chains of thought, which are later summarized into a final answer. % This preserves the reasoning fidelity of the base LLM and enables efficient scaling without continual pretraining or structural modifications. 
Other works \citep{yang2025multiverse, pan2025learningadaptiveparallelreasoning, rodionov2025hogwildinferenceparallelllm, jin2025learningpromisescalinglanguage} propose to automatically identify subtasks that can be solved in parallel. While effective for compositional tasks, it relies on explicit subtask decomposition, and these works focus on efficiency rather than accuracy improvement.
\name, on the other hand, does not assume any subtask structure and improves both efficiency and accuracy by mitigating single-path failure cases (\eg hallucinations or local optima) via diversity in reasoning. By integrating multiple KV caches in a summarization stage, our method scales inference without sacrificing correctness or requiring verifier models.
% Another line of work focuses on replacing common autoregressive LLMs by modifying model architecture. Diffusion language models parallelly sample multiple tokens in each diffusion step \citep{he-etal-2023-diffusionbert, ye2023diffusionllm, zhao2025d1, wang2025remasking, arriola2025block}. However, recent work \citep{feng2025theoreticalbenefitlimitationdiffusion} has found that for some reasoning chains the sampling steps could scale linearly with sequence length, thereby eliminating the efficiency advantage. Besides, the performance of these non-autoregressive language models is still not able to scale to complex reasoning tasks. 
% PARSCALE \citep{chen2025parallelscalinglawlanguage} explore parallel scaling by applying learnable transformations to the input and duplicate it, execute forward passes of duplicated inputs and aggregating them at each output token. 
% Our method, on the other hand, independently generate several reasoning paths for each question instead of aggregating at each token, eliminating the need of large-scale continual pre-training and can preserve the reasoning capability of original LLM largely. 
% Our concurrent work \citep{yang2025multiverse} can automatically detect subtasks that can be divided into parallel execution, which can solve problems more efficiently. However, this method could only work for problems that can be divided into subtasks and doesn't improve the accuracy of LLMs. 
% Our work, on the other hand, doesn't rely on subtasks and can improve LLM's reasoning capacity by decreasing thinking traps of single-sequence reasoning LLMs. 

%% file: tex/conclusion.tex
\section{Conclusion}

Our work identifies a fundamental limitation of current test-time compute strategies, where sequential reasoning gets caught in Tunnel Vision set by its own initial tokens. 
We demonstrate that this is not an inherent model capability limit but a failure of the scaling strategy itself. 
We then introduce \name, a framework for native parallel reasoning that sidesteps these traps by generating and synthesizing multiple thought-paths simultaneously. 
This approach proves that scaling compute in parallel (width) is a more effective and efficient path to superior reasoning than simply scaling sequentially (depth). 
While our method is a significant first step, future work could explore more advanced aggregation strategies and deeper reinforcement learning. 
Fundamentally, our results argue that to unlock the next level of reasoning, LLM test-time compute should scale in parallel.

%% file: tex/appendix.tex
\appendix
\section{Appendix}

% \subsection{Prompt used for generating SFT data}
% \label{sec:prompt}

% To generate the supervised fine-tuning (SFT) dataset, we employ the following prompt to sample answers from the teacher model (DeepSeek-R1 \citep{deepseekr1}) with a temperature of \xx.  
% For each question, this prompt is sent to the same teacher model \xx times, yielding \xx distinct answers. 

% \begin{AIbox}{Data Generation Prompt}
% \end{AIbox}

% \subsection{Similarities of thought embeddings}
\subsection{Supervised Fine-Tuning (SFT) Details}
\label{appendix:train_settings}
% Providing details of SFT training configuration
This section details the configuration used for supervised fine-tuning (SFT) of the large language model.

\begin{table}[h]
\centering
\begin{tabular}{lc}
\toprule
\textbf{Parameter} & \textbf{Value} \\
\midrule
Batch Size & 1 \\
Gradient Accumulation Steps & 8 \\
Learning Rate & $1 \times 10^{-5}$ \\
Training Epochs & 3 \\
Context Length & 28,672 \\
Hardware & 4 GPUs \\
Learning Rate Scheduler & Constant \\
Warmup Ratio & 0.1 \\
Weight Decay & 0.05 \\
Max Gradient Norm & 0.5 \\
\bottomrule
\end{tabular}
\caption{SFT Training Configuration for DeepSeek-R1-Distill-Qwen-1.5B}
\end{table}

\begin{table}[h]
\centering
\begin{tabular}{lc}
\toprule
\textbf{Parameter} & \textbf{Value} \\
\midrule
Batch Size & 1 \\
Gradient Accumulation Steps & 4 \\
Learning Rate & $2 \times 10^{-5}$ \\
Training Epochs & 2 \\
Context Length & 28,672 \\
Hardware & 8 GPUs \\
Learning Rate Scheduler & Cosine with Minimum LR \\
Warmup Ratio & 0.1 \\
Weight Decay & 0.05 \\
Max Gradient Norm & 0.5 \\
\bottomrule
\end{tabular}
\caption{SFT Training Configuration for DeepSeek-R1-Distill-Qwen-7B}
\end{table}

\newpage

\subsection{Examples of training data}
\label{appendix:sft_data}

After sampling multiple answers from the teacher model, we construct the SFT training data as follows:  
(1) concatenate all sampled answers as separate reasoning paths;  
(2) insert special tokens to delineate the parallel reasoning paths;  
(3) append the ground-truth final answer. 
An example is shown below: 

\begin{AIbox}{SFT Data Example}
\textless|User|\textgreater \emph{Determine the number of solutions to the equation $z^{100} = 1 + i$ that lie in the third quadrant of the complex plane.} You FIRST think about the reasoning process as an internal monologue and then summarize the reasoning process to get the final answer. The summary process MUST BE enclosed within \textless summary\textgreater\ \textless/summary\textgreater\ tags. The final answer MUST BE put in \texttt{\textbackslash boxed\{\}}.\textless|Assistant|\textgreater\textless think\textgreater\textless vllm\_pad\textgreater \vspace{1em}

\textless think1\textgreater

I need to solve $z^{100} = 1 + i$ and find how many solutions are in the third quadrant of the complex plane. First, I recall that complex numbers can be represented in polar form. So, I'll start by writing $1 + i$ in polar form ... \vspace{1em}

\textless think2\textgreater

Okay, so I need to find how many solutions of the equation $z^{100} = 1 + i$ are in the third quadrant. Hmm, let's start by recalling that complex equations like this one can often be solved using polar form and De Moivre's theorem. Let me jot down what I remember ...

\textless/think2\textgreater \vspace{1em}

\textless think3\textgreater

Okay, so I need to figure out how many solutions of the equation \( z^{100} = 1 + i \) lie in the third quadrant of the complex plane. Hmm, let's start by recalling that in the complex plane, the third quadrant is where both the real and imaginary parts of the complex number are negative. So any solution \( z \) must satisfy \( \text{Re}(z) < 0 \) and \( \text{Im}(z) < 0 \) ... \vspace{1em} 

\textless think4\textgreater

I need to solve the equation \( z^{100} = 1 + i \) and find how many solutions are in the third quadrant of the complex plane. First, I recall that complex numbers can be expressed in polar form, which might help with exponents.
So, let \( z = re^{i\theta} \), where r is the modulus and $\theta$ is the argument. Then \( z^{100} = r^{100} e^{i 100 \theta} \) ...

\textless/think4\textgreater \vspace{1em}

\textless summary\textgreater

By analyzing multiple reasoning processes above, I concluded that: The final answer is \(\boxed{25}\).

\textless/summary\textgreater

\end{AIbox}

\subsection{Examples of parallel answers}

We illustrate the effectiveness of \name through an example.  
Question: ``\textit{Jen enters a lottery by picking $4$ distinct numbers from $S=\{1,2,3,\cdots,9,10\}.$ $4$ numbers are randomly chosen from $S.$ She wins a prize if at least two of her numbers were $2$ of the randomly chosen numbers, and wins the grand prize if all four of her numbers were the randomly chosen numbers. The probability of her winning the grand prize given that she won a prize is $\tfrac{m}{n}$ where $m$ and $n$ are relatively prime positive integers. Find $m+n$.} ''
\name produces the following output (shortened for brevity):

\begin{AIbox}{Answer Example}
\textless think1\textgreater

Use the conditional probability formula
\[
P(A|B) = \frac{P(A \cap B)}{P(B)},
\]
where $A$ is winning the grand prize (all four numbers correct), and $B$ is winning a prize (at least two numbers correct).  
Calculate
\[
P(A) = \frac{1}{\binom{10}{4}} = \frac{1}{210}.
\]
Then calculate $P(B)$ as the probability of having at least two numbers correct, using the complement method:
\[
P(B) = 1 - P(0) - P(1) 
= 1 - \frac{\binom{6}{4}}{210} - \frac{\binom{4}{1}\binom{6}{3}}{210}
= \frac{115}{210} = \frac{23}{42}.
\]
The final conditional probability is
\[
\frac{\frac{1}{210}}{\frac{23}{42}} = \frac{1}{115},
\]
hence $m + n = 116$. 

\textless/think1\textgreater \vspace{1em}

\textless think2\textgreater

calculate $P(B)$ directly as the probability of having at least two numbers correct:
\[
\binom{4}{2}\binom{6}{2} + \binom{4}{3}\binom{6}{1} + \binom{4}{4}\binom{6}{0} = 115.
\]
The conditional probability is $\frac{1}{115}$ \vspace{1em}

\textless think3\textgreater

Directly compute the intersection to confirm:
\[
P(A|B) = \frac{1}{115}.
\] \vspace{1em}

\textless think4\textgreater

Use the hypergeometric distribution formula to calculate $P(B)$:
\[
P(B) = \sum_{k=2}^4 \frac{\binom{4}{k}\binom{6}{4-k}}{\binom{10}{4}} = \frac{115}{210},
\] \vspace{1em}

\textless summary\textgreater

By analyzing multiple reasoning processes above, I concluded that: The final answer is \(\boxed{116}\).

\textless/summary\textgreater
\end{AIbox}

This example demonstrates that the generated parallel answers employ diverse reasoning strategies that complement each other. 
By considering multiple reasoning paths, \name can mitigate the risk of being misled by a flawed initial line of thought.